\pdfoutput=1

\documentclass[11pt]{article}

\usepackage{emnlp2021}

\usepackage{times}
\usepackage{latexsym}

\usepackage[T1]{fontenc}

\usepackage[utf8]{inputenc}

\usepackage{microtype}

\usepackage{hyperref}
\usepackage{graphicx}
\usepackage{multirow}
\usepackage{appendix}
\usepackage{wrapfig,lipsum,booktabs}
\usepackage{color}
\usepackage{url}
\usepackage{subcaption}

\usepackage{amsmath,amsfonts,bm}

\def\eqref#1{equation~\ref{#1}}

\def\1{\bm{1}}

\DeclareMathAlphabet{\mathsfit}{\encodingdefault}{\sfdefault}{m}{sl}
\SetMathAlphabet{\mathsfit}{bold}{\encodingdefault}{\sfdefault}{bx}{n}

\title{Mathematical Word Problem Generation from Commonsense Knowledge Graph and Equations}

\author{Tianqiao Liu, Qiang Fang, Wenbiao Ding, Hang Li, Zhongqin Wu, Zitao Liu\thanks{\quad The corresponding author: Zitao Liu} \\
        TAL Education Group, Beijing, China \\
        \texttt{\{liutianqiao,fangqiang,dingwenbiao,lihang4,wuzhongqin,liuzitao\}@tal.com} \\}
\begin{document}
\maketitle

\begin{abstract}
There is an increasing interest in the use of mathematical word problem (MWP) generation in educational assessment. Different from standard natural question generation, MWP generation needs to maintain the underlying mathematical operations between quantities and variables, while at the same time ensuring the relevance between the output and the given topic. To address above problem, we develop an end-to-end neural model to generate diverse MWPs in real-world scenarios from commonsense knowledge graph and equations. The proposed model (1) learns both representations from edge-enhanced Levi graphs of symbolic equations and commonsense knowledge; (2) automatically fuses equation and commonsense knowledge information via a self-planning module when generating the MWPs. Experiments on an educational gold-standard set and a large-scale generated MWP set show that our approach is superior on the MWP generation task, and it outperforms the SOTA models in terms of both automatic evaluation metrics, i.e., BLEU-4, ROUGE-L, Self-BLEU, and human evaluation metrics, i.e., equation relevance, topic relevance, and language coherence. To encourage reproducible results, we make our code and MWP dataset public available at \url{https://github.com/tal-ai/MaKE_EMNLP2021}.
\end{abstract}

\section{Introduction}
\label{sec:intro}
A mathematical word problem (MWP) is a coherent narrative that provides clues to the underlying correct mathematical equations and operations between variables and numerical quantities \citep{cetintas2010joint,moyer1984story}. MWPs challenge a student from a wide range of skills such as literacy skills for understanding the question, analytical skills for recognizing the problem type and applying arithmetical operators \citep{rembert2019exploring,moon2019illmatics}. Table \ref{tab:example} shows one such problem\footnote{One MWP example from \url{https://www.hackmath.net/en/math-problem/56}} where students are asked to infer the counts of chickens and rabbits.


\begin{table}[!bhpt]
\small
\vspace{-0.2cm}
\begin{center}
\scalebox{0.9}{
\begin{tabular}{@{}l|l|l|l@{}}
\toprule
Math Word Problem          & \multicolumn{3}{l}{\begin{tabular}[c]{@{}l@{}}Chickens and rabbits were in the yard. \\ Together they had 27 heads and 86 legs. \\ How many rabbits were in the yard?\end{tabular}} \\ \midrule
\multirow{2}{*}{Equations} & x+y=27                                                      & \multirow{2}{*}{Solutions}                                                    & x=11                                                   \\
							& 2x+4y=86                                                    &                                                                               & y=16                                                   \\ \bottomrule
\end{tabular}
}
\caption{An illustrative example of an MWP.} \label{tab:example}
\end{center}
\vspace{-0.5cm}
\end{table}



In this paper, our objective is to automatically generate well-formed MWPs. Such automation will not only reduce the teachers' burden of manually designing MWPs, but provide students with a sufficiently large number of practice exercises, which help students avoid rote memorization \citep{williams2011generating,wang2016dimensionally}.

A large spectrum of models have been developed and successfully applied in a broad area of natural question generation (NQG) \citep{pan2019recent,li2018visual,liu2020personalized,sun2018answer,zhang2019addressing,kurdi2020systematic,guan2021long,guan2021openmeva} and there has been a recent movement from the NQG community towards automatic generation of MWPs \citep{koncel2016theme,polozov2015personalized,zhou2019towards}. For example, \citet{koncel2016theme} proposed a two-stage rewriting approach to edit existing human-authored MWPs. \citet{polozov2015personalized} conducted the MWP generation as a constrained synthesis of labeled logical graphs that represent abstract plots. 

In general, there exists a large number of NQG models representing various text data and their syntax and semantics \citep{pan2019recent}. However, automatic generation of MWPs still presents numerous challenges that come from special characteristics of real-world educational scenarios as follows: (1) MWP generation models need to not only generate fluent sentences but understand the mathematical variables, numerical quantities, operations, and their relations. Moreover, the models are supposed to be able to generalize to unseen equations. (2) Multiple studies have found that MWPs with real-life plots help conceptual knowledge understanding, discourse comprehension and children engagement \citep{carpenter1980solving,rembert2019exploring}. (3) Computerized educational assessment systems require diverse MWP results even given similar input equations, which helps prevent students from rote memorization \citep{deane2003automatic}.

To overcome the above challenges, in this paper, we present a novel neural generation model \textbf{MaKE} (short for \textbf{Ma}thematical word problem generation from commonsense \textbf{K}nowledge and \textbf{E}quations), which aims to automatically generate coherent and diverse MWPs from given equations in students' real-life scenarios. More specifically, to fully understand the mathematical variables, numerical quantities, operations, and their relations, equations are transformed into an edge-enhanced Levi graph. We adopt the gated graph neural networks (GGNNs) to learn representative embeddings from the equation based symbolic Levi graph. Meanwhile, the same procedure is applied to the external commonsense based knowledge graph (CSKG), which helps generate topic-relevant and semantically valid sentences in real-life settings. We choose to use the conditional variational autoencoder (VAE) framework to generate MWPs from diversity promoting latent states. Furthermore, in the decoding stage, we develop a self-planning module to dynamically select and fuse information from both equations and commonsense knowledge, which improves syntax structure of generated MWP sentences. Overall this paper makes the following contributions:

\begin{itemize}
\setlength{\itemsep}{0.1pt}
\item We propose a GGNN based conditional VAE model for MWP generation. To the best of our knowledge, we are the first to introduce the combinational architecture of GGNN and condition VAE for MWP generation.
\item We design a novel self-planning decoding module to wisely fuse information from equations and commonsense knowledge with implicit schedule, which helps generate semantically valid MWPs.
\item The proposed model achieves the SOTA scores and outperforms existing methods by a significant margin on real-world educational MWP datasets from both automatic machinery and human evaluation metrics.
\end{itemize}

\section{Related Work}
\label{sec:related}
\subsection{Natural Question Generation}

Previous research has directly approached the task of automatically generating questions for many useful applications such as augmenting data for the QA tasks \citep{li2018visual,sun2018answer,zhang2019addressing}, helping semantic parsing \citep{guo2018question} and machine reading comprehension \citep{yu2020generating,yuan2017machine}, improving conversation quality \citep{mostafazadeh2016generating,dong2019unified}, and providing student exercises for education purposes \citep{koncel2016theme}.


Various NQG methods are developed which can be divided into two categories: heuristic based approaches and neural network based approaches \citep{pan2019recent,kurdi2020systematic}. The former generates questions in two stages: it first obtains intermediate symbolic representations and then constructs the natural language questions by either rearranging the surface form of the input sentence or generating with pre-defined question templates. The latter neural approaches view the NQG task as a sequence-to-sequence (seq2seq) learning problem and jointly learn generation process in an end-to-end manner \citep{yao2018teaching, zhou2018sequential}. 

\subsection{Math Word Problem Generation}

Different from standard NQG tasks, generating MWPs not only needs the syntax, semantics and coherence of the output narratives, but requires understandings of the underlying symbolic representations and the arithmetic relationship between quantities. In general, MWP generation approaches can be divided into three categories: (1) template based approaches; (2) rewriting based approaches; and (3) neural network based approaches. 

Template based approaches usually fall into a similar two-stage process: they first generalize an existing problem into a template or a skeleton, and then generate the MWP sentences from the templates \citep{,williams2011generating,polozov2015personalized,bekele2020automatic}. \citet{deane2003automatic} used semantic frames to capture both scene stereotypical expectations and semantic relationships among words and utilized a variant of second-order predicate logic to generate MWPs. \citet{wang2016dimensionally} leveraged the binary expression tree to represent the story of the MWP narrative and composed the natural language story recursively via a bottom-up tree traversal. Template based approaches heavily rely on the tedious and limited hand-crafted templates, leading to very similar generated results. This cannot meet the demand of a large number of high-quality and diverse MWPs.

Rewriting based approaches target the MWP generation problem by editing existing human-written MWP sentences to change their theme without changing the underlying story \citep{koncel2016theme,moon2019illmatics}. For example, \citet{koncel2016theme} proposed a rewriting algorithm to construct new texts by substituting thematically appropriate words and phrases. Rewriting based approaches are more flexible compared with templates based approaches. However, there are several drawbacks that prevent them from providing the large number of MWPs. First, the generation process is based on existing MWPs, which significantly limits the generation ability. Second, students easily fall into rote memorization since it is too trivial to notice that the underlying mathematical equations are still unchanged. 

Recent attempts have been focused on exploiting neural network based approaches that generating MWPs from equations and topics in an end-to-end manner \citep{zhou2019towards,liyanage2020multi}. \citet{zhou2019towards} designed a neural network with two encoders to fuse information of both equations and topics and dual-attention mechanism to generate relevant MWPs. \citet{liyanage2020multi} tackled the generation problem by using the long short term memory network with enhanced input features, such as character embeddings, word embeddings and part-of-speech tag embeddings. 

The closest work to our approach is \citet{zhou2019towards} and the main differences are as follows: (1) \citet{zhou2019towards} directly encode the equation by a single-layer bidirectional gated recurrent unit (GRU), while we first convert equations into Levi graph and conduct the encoding by the GGNN model; (2) instead of directly using the pre-trained embeddings of similar words given the topic, we choose to learn the topic relevant representations from an external CSKG; and (3) we choose to use the VAE framework to promoting more diverse results.

\section{Learning from Commonsense Knowledge and Equations}
\label{sec:method}
Our objective is to automatically generate a significant number of diverse MWPs in students' real-life scenarios from valid equations. In addition, we support the personalized generation in which students (or teachers) can determine the story plots of MWPs by specifying topics and mapping relations between variables and entities (i.e., ``x: chicken, y: rabbits'', ``x: apple, y: banana'', etc.). A topic indicates a type of real-world scenarios, such as animals, fruits, etc.

\begin{figure*}[!hbpt]
    \vspace{-0.5cm}
    \begin{center}
    \includegraphics[width=\textwidth]{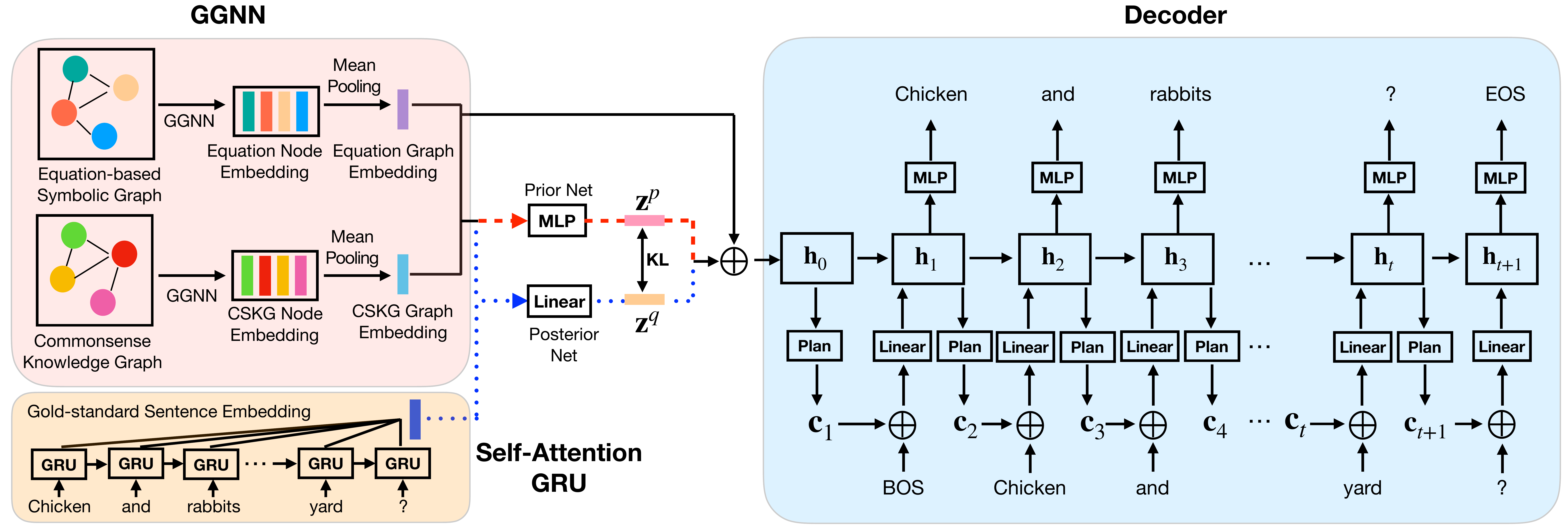}
    \end{center}
    \caption{The overview of the proposed framework. The blue dot line (or red dash line) is only enabled in the training (or inference) stage. $\oplus$ denotes the vector concatenation. \textbf{Linear} represents the linear transformation and \textbf{Plan} denotes the self-planning module (discussed in Section \ref{sec:cvae}).}
    \label{fig:model}
    \vspace{-0.7cm}
\end{figure*}

As shown in Figure \ref{fig:model}, we adopt the encoder-decoder generation framework. The input includes a set of equations and a knowledge graph with a specific topic. We construct Levi graphs \citep{levi1942finite} from symbolic equations and the CSKG respectively  (See Section \ref{sec:levi}). After that, we employ GGNNs to extract the full graph structure information about equations and real-life story plots (See Section \ref{sec:ggnn}). Then, we generate target sentence by a conditional VAE with a self-planning module (See Section \ref{sec:cvae}). The self-planning module enables the decoder to pay different portions of attention to the equations and the CSKG. 

Please note that in this paper, we focus on generating MWPs with linear equations of two variables without any constraint. Our framework can be easily generalized into MWPs with different numbers of variables with little modification.

\subsection{Levi Graph Construction}
\label{sec:levi}
\subsubsection{Equation Based Symbolic Graph}
\label{sec:equation}

The equation based symbolic graph is designed to capture the relations among mathematical variables and numerical quantities, and build connections between mathematical variables and the corresponding commonsense knowledge. In this work, we consider the linear equations (with two variables) behind the MWPs as $ax + by = m; cx + dy = n$, where $x$ and $y$ are the variables and $a$, $b$, $c$, $d$, $m$, and $n$ are positive integer quantities. More equation variants are discussed in Appendix \ref{sec:equation_variants}.

Equations are first converted to a symbolic graph as shown in Figure \ref{fig:dual_graph} (a). In the symbolic graph, edge labels, i.e., \emph{Add to res}, \emph{Mul}, etc. representing the mathematical relations play important roles in the MWP generation, where ``\emph{Add to res}'' indicates addition operation to the result operand and ``\emph{Mul}'' indicates multiply operation. In order to well capture such relations, we model the edge labels as explicit nodes. Following previous work in \citet{beck2018graph}, we transform the symbolic graph into its equivalent edge-enhanced Levi graph \citep{levi1942finite} by adding two nodes for each labeled edge. One node denotes the forward direction of the relation and one represents the reverse. By adding reverse nodes, we encourage more information flow from the reverse direction, in the same way, RNN-based encoders benefit from right-to-left propagation. Furthermore, we explicitly add self-loop edges to each node in the Levi graph. The symbolic Levi graph is depicted in Figure \ref{fig:dual_graph} (b). More details on Levi graph transformation can be found in Appendix \ref{sec:levi_transformation}.

\subsubsection{Commonsense Based Knowledge Graph}

\begin{figure*}[!bpth]
    \vspace{-0.5cm}
    \center
    \includegraphics[width=\textwidth]{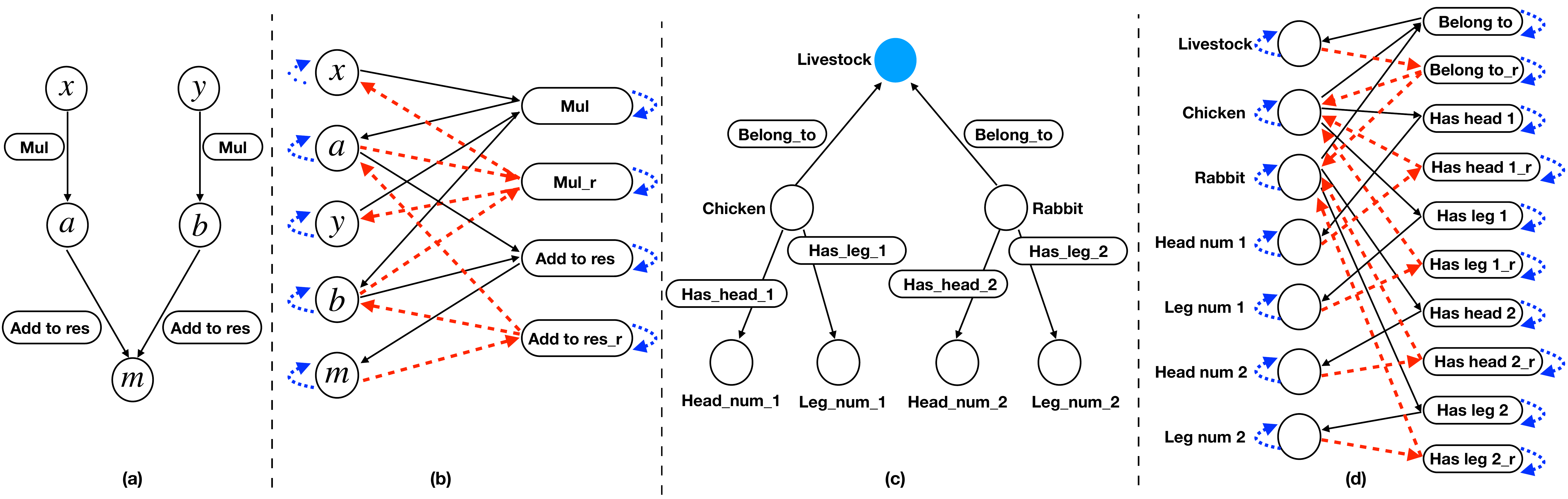}
    \caption{(a) a sample symbolic graph of equation ``ax+by = m''; (b) the edge-enhanced Levi graph of the same equation; (c) an illustrative sample of the CSKG under topic \emph{livestock}; (d) the corresponding edge-enhanced Levi graph of CSKG. The red dash arrows represent the reverse edges and the blue dot arrows represent the self-loop edges. The subscript ``\_r'' denotes the artificially added reverse nodes.The blue node in Figure \ref{fig:dual_graph}(c) denotes the given topic.}
    \label{fig:dual_graph}
    \vspace{-0.5cm}
\end{figure*}

In order to generate valid questions in students' real-life scenarios, we utilize explicit knowledge from a self-derived CSKG specifically designed for MWP generation. We have to admit that our CSKG is of particularly tiny size compared to publicly available knowledge graphs like ConceptNet and Wikipedia. However, we have exclusive relationships that can be utilized for MWPs generations, i.e., (apple, has unit of measurement, pound), (banana, has price unit, yuan), (chicken, has feet number, 2), etc. These commonsense knowledge triples are extracted from MWP texts in a semi-automatic manner. Specifically, for each MWP, we first apply the part-of-speech tagger from Stanford CoreNLP\footnote{\url{https://nlp.stanford.edu/software/tagger.shtml}} with some heuristic rules for automatic commonsense knowledge extraction. Furthermore, because the generation process requires high-quality commonsense information, we ask crowd workers to verify the extracted results, which includes both entities and the corresponding attributes. For example, for the MWP shown in Table \ref{tab:example}, the auto-parsed entities are \emph{Chickens} and \emph{Rabbits} and the extracted relations are (1) a \emph{belong\_to} relation showing that the \emph{Chickens} belong to \emph{livestock}; and (2) a \emph{has\_head\_entity} relation showing that the \emph{Chickens} have head entity \emph{head}. The similar relations about Rabbits are extracted as well. We form these triples into our commonsense knowledge graphs. Moreover, we explicitly select a few entities from the above extraction process as ``topics'' and these topic terms can be revised by the crowd workers if they are mis-extracted. The topic entities are obtained from a given K-12 educational vocabulary. Figure \ref{fig:dual_graph} (c) illustrates a sample of a CSKG with a topic of \emph{Livestock}.

With the help of CSKG, students or teachers are able to set their own preferences when generating MWPs by choosing different topics, such as zoo, transportation, etc. This external commonsense knowledge provides additional background information that improves the generated results diversity. Moreover, the CSKG improves the generation quality by alleviating ill-informed wordings or sentences. For instance, in spite of no grammatical errors, it makes no sense to have ``rabbits live in the ocean'' or ``apple has two feet''. Similar to the Levi graph construction procedure in Section \ref{sec:equation}, we introduce additional nodes for relations in CSKG and add reverse and self-loop edges. The CSKG Levi graph is shown in Figure \ref{fig:dual_graph} (d).

\subsection{Gated Graph Neural Encoding}
\label{sec:ggnn}

Following the success of GGNN models \citep{beck2018graph,ruiz2019gated}, we use GGNNs to capture both the mathematical relations among variables and quantities and the real-life associations among entities in the MWPs. Specifically, let $\mathcal{G} = \{\mathcal{V}, \mathcal{E}\}$ be an edge-enhanced Levi graph where $\mathcal{V}$ and $\mathcal{E}$ are the sets of nodes and edges. Let $a^{v,u}$ be the similarity between node $v$ and node $u$ from its row-wise normalized adjacent matrix. Given an input Levi graph $\mathcal{G}$ that may represent either the equations or the CSKG, the basic recurrence of the GGNN model is defined as follows:

\vspace{-0.5cm}
\begin{small}
\begin{align*}
\mathbf{g}_0^v &= \mathbf{e}_0^v; \quad \bm{\gamma}^v_t = \sum_{u \in N_{(v)}}a^{v,u} \mathbf{g}_{t-1}^{u}; \\
\mathbf{z}_t^v &= \sigma(\mathbf{W}^z \bm{\gamma}^v_t + \mathbf{U}^z \mathbf{g}_{t-1}^v); \\ 
\mathbf{r}_t^v &= \sigma(\mathbf{W}^r \bm{\gamma}^v_t + \mathbf{U}^r \mathbf{g}_{t-1}^v);\\ 
\widetilde{\mathbf{g}_t^v} &= \tanh(\mathbf{W}^h \bm{\gamma}^v_t + \mathbf{U}^h(\mathbf{r}_t^v \odot \mathbf{g}_{t-1}^v));\\
\mathbf{g}_t^v &= (1-\mathbf{z}_t^v) \odot \mathbf{g}_{t-1}^v + \mathbf{z}_t^v \odot \widetilde{\mathbf{g}_t^v}
\end{align*}
\end{small}
\vspace{-0.5cm}

\noindent where $\mathbf{e}_0^v$ denotes the initial embedding of node $v$. $N_{(v)}$ is the set of neighbor nodes for $v$ and $\sigma$ is the sigmoid function. $\odot$ is the component-wise multiplication function and $\mathbf{z}_t^v$ and $\mathbf{r}_t^v$ are gating vectors. 

Let $\mathbf{G}_0 = [\mathbf{g}_0^1; \mathbf{g}_0^2; \cdots; \mathbf{g}_0^{|\mathcal{V}|} ]$ be the initial word embedding matrix of all the nodes and $\mathbf{G}_n$ be the matrix of representation of node embeddings from the above GGNN model after \emph{n} iterations, i.e., $\mathbf{G}_n = [\mathbf{g}_n^1; \mathbf{g}_n^2; \cdots, \mathbf{g}_n^{|\mathcal{V}|} ]$. Similar to \citet{he2016deep}, we ease the downstream learning tasks with embedding augmentation. We apply a linear transformation on the concatenation of $\mathbf{G}_0$ and $\mathbf{G}_n$, i.e., $\mathbf{G}_* = \mathbf{W}_* [\mathbf{G}_0; \mathbf{G}_n]$. Such augmented node representations contain abstract context information, which are used in our language generator in Section \ref{sec:cvae}. Let $\mathbf{G}_*^e$ and $\mathbf{G}_*^k$ be the augmented GGNN embeddings of the equations and the CSKG. Meanwhile we apply a mean pooling operation over $\mathbf{G}_*^e$ and $\mathbf{G}_*^k$ to get the graph-level equation representation ($\mathbf{g}_*^e$) and CSKG representation ($\mathbf{g}_*^k$).

\subsection{Conditional VAE with Self-Planning}
\label{sec:cvae}

In this section, we introduce our VAE architecture with the self-planning module for the MWP generation. Our self-planning module makes dynamic fusion on the learned representations of equations and CSKG to generate the MWPs.

Let $Y$ be the random variable representing the texts of MWPs and $Z$ be the diversity promoting latent variable of the distribution of the MWPs. Let $C$ be the random variable representing the conditions of both the explicit equations and the implicit CSKG learned from GGNNs. We model the MWP generation by the conditional distribution as follows: $p(Y|C)= \int p(Y|C, Z)p(Z|C) dZ$ where $p(Y|C, Z)$ is the MWP generator and $p(Z|C)$ is the prior net. Since the integration of $Z$ is intractable, we apply variational inference and optimize the evidence lower bound as follows:


\vspace{-0.6cm}

\begin{small}
\begin{equation*} 
\begin{split}
\log p(Y|C) & \ge \mathbb{E}_{q(Z|C, Y)}\Big[\log p(Y|C, Z)\Big] \\
& - D_{KL}\Big(q(Z|C, Y) || p(Z|C)\Big)
\end{split}
\end{equation*}
\vspace{-0.4cm}
\end{small}


\noindent where $D_{KL}(\cdot || \cdot)$ denotes the KL-divergence. 

Following conventions, we assume both the prior net and posterior net of $Z$ following the isotropic Gaussian distributions, i.e., $p(Z|C) \sim \mathcal{N}(\boldsymbol{\mu}^p, \sigma^p \mathbf{I})$ and $q(Z|C, Y) \sim \mathcal{N}(\boldsymbol{\mu}^q, \sigma^q \mathbf{I})$. The prior net only encodes the given conditions of both the explicit equations and the implicit CSKG while the posterior net encodes both given conditions and the texts of MWPs. Both the prior net and the posterior net are built upon the GGNNs shown in Figure \ref{fig:model} as follows:

\begin{small}
\vspace{-0.5cm}
\begin{align*}
[\boldsymbol{\mu}^p; \log \boldsymbol{\sigma}^p] &= \textbf{MLP}\big([\mathbf{g}_*^e; \mathbf{g}_*^k]\big); \\
[\boldsymbol{\mu}^q; \log \boldsymbol{\sigma}^q] &= \mathbf{W}^q \big([\mathbf{g}_*^e; \mathbf{g}_*^k; \textbf{GRU}(\mathbf{y})]\big) + \mathbf{b}^q \nonumber
\end{align*}
\vspace{-0.5cm}
\end{small}

Due to the flexibility of language, there may exist more than one reasonable expression that covers the same input but in different sequence. For example, ``Chickens and rabbits  were in the yard. Together they had 27 heads and 86 legs." can be rewritten as ``Teacher finds 27 heads and 86 legs in the yard, in which there are only chickens and rabbits.". The former expression can be viewed as the plan of first generating commonsense sentences and then the symbolic sentences, while the latter one is first generating symbolic sentences then commonsense sentences. We capture such diversity of reasonable presentations with both latent variable $Z$ and input graphs $C$. Different samples of $Z$ will lead to different self-planning results. To start the decoding process, we initialize the hidden state $\mathbf{h}_0 = [\mathbf{z}; \mathbf{g}_*^e; \mathbf{g}_*^k]$, where $\mathbf{z}$ is sampled from the posterior net $q(Z|C, Y) \sim \mathcal{N}(\boldsymbol{\mu}^q, \boldsymbol{\sigma}^q \mathbf{I})$ and the prior net $p(Z|C) \sim \mathcal{N}(\boldsymbol{\mu}^p, \boldsymbol{\sigma}^p \mathbf{I})$ during the training and inference procedures respectively. 


At each decoding time step $t$, we dynamically decide the portions of input information from the equations and the CSKG respectively based on the current hidden state $\mathbf{h}_t$, which can keep track of the current generating state. We use the attention mechanism to conduct the self-planning between explicit symbolic equations and implicit CSKG. The dynamic self-planning module takes the decoder's current hidden state ($\mathbf{h}_t$), node representations of equations ($\mathbf{G}_*^e$) and CSKG ($\mathbf{G}_*^k$) as input and outputs the context-aware planning state ($\mathbf{c}_t$) of the current time step. Specifically, we compute $\mathbf{c}_t$ as follows:

\begin{small}
\vspace{-0.5cm}
\begin{align*}
\mathbf{c}_t &= \beta_t * \mathbf{c}_t^e + (1 - \beta_t) * \mathbf{c}_t^k; \quad \beta_t = softmax(\mathbf{W}^{\beta} \mathbf{h}_t); \\
\mathbf{c}_t^e &= \sum_{v \in \mathcal{V}^e} \alpha_{t,v}^e \mathbf{g}_v^e; \quad \mathbf{c}_t^k = \sum_{v \in \mathcal{V}^k} \alpha_{t,v}^k \mathbf{g}_v^k; \\
\alpha_{t,v}^e &= \exp(\mathbf{o}^e_{t,v})/\sum_{v' \in \mathcal{V}^e} \exp(\mathbf{o}^e_{t,v'}); \\ 
\alpha_{t,v}^k &= \exp(\mathbf{o}^k_{t,v}) / \sum_{v' \in \mathcal{V}^k} \exp(\mathbf{o}^k_{t,v'});  \\
\mathbf{o}^e_{t,v} &= {\mathbf{v}^e}^\top \tanh (\mathbf{W}^e \mathbf{h}_t + \mathbf{U}^e \mathbf{g}^e_v); \\
\mathbf{o}^k_{t,v} &= {\mathbf{v}^k}^\top \tanh (\mathbf{W}^k \mathbf{h}_t + \mathbf{U}^k \mathbf{g}^k_v)
\end{align*}
\end{small}

\vspace{-0.5cm}

\noindent where $\beta_t$ represents the self-planning distribution at time step $t$.

The final context vector is the fusion of the symbolic and commonsense knowledge graphs. The next-step hidden state ($\mathbf{h}_{t+1}$) is the combination of current hidden state ($\mathbf{h}_{t}$), self-planning context state ($\mathbf{c}_t$) and the representation of currently generated word ($\mathbf{w}_t$), i.e, $\mathbf{h}_{t+1} = \textbf{GRU}(\mathbf{h}_t, \mathbf{W}^d[\mathbf{c}_t; \mathbf{w}_t] + \mathbf{b}^d)$ where $\mathbf{W}^d$ and $\mathbf{b}^d$ are the linear transformation matrix and the bias term. We further generate the next word by feeding hidden state $\mathbf{h}_{t+1}$ to linear transformation and softmax layer to get the next-token probability distribution.



The final objective function consists (1) maximizing the probability of ground-truth sequence texts, which promotes the predictions generated by the posterior net and the MWP generator closer to the distribution of the gold-standard data; and (2) minimizing the KL-divergence between posterior distribution ($p(Z|C, Y)$) and prior distribution ($p(Z|C)$).

\section{Experiments}
\label{sec:exp}
\begin{table*}[!htbp]
\footnotesize
\begin{center}
\begin{tabular}{@{}lccccccc@{}} \toprule
Method & BLEU-4 & METEOR & ROUGE-L & Self-BLEU & Equation Rel. & Topic Rel. & Language Coh.\\ 
\midrule
Template & 19.836 & 22.685 & 36.545 & 57.309 & 2.493 (0.377) & 2.276 (0.419) & 2.256 (0.431)\\
CVAE & 20.971 & 23.278 & 37.557 & 57.560 & 1.880 (0.514) & 2.266 (0.498) & 1.819 (0.529)\\
MAGNET & 15.075 & 20.084 & 35.695 & 71.032 & 1.850 (0.579) & 2.633 (0.502) & 1.725 (0.502)\\
UniLM & 17.499 & 20.562 & 34.934 & 74.198 & 2.103 (0.614) & 2.546 (0.720) & 1.683 (0.605)\\
Transformer & 21.612 & 24.423 & 39.250 & 77.571 & 2.486 (0.537) & 2.580 (0.413) & 2.313 (0.476)\\
\midrule
MaKE w/o symbolic & 19.788 & 22.386 & 38.459 & 63.849 & 2.016 (0.670) & \textbf{2.890} (0.317) & 2.376 (0.522)\\
MaKE w/o CSKG & 15.441 & 19.662 & 33.188 & \textbf{56.683} & 2.491 (0.682) & 2.240 (0.271) & 2.092 (0.634)\\
MaKE w/o planning & 21.611 & 23.782 & 40.053 & 62.549 & 2.575 (0.695) & 2.695 (0.528) & 2.195 (0.472)\\
MaKE & \textbf{23.322} & \textbf{24.537} & \textbf{40.076} & 60.29 & \textbf{2.668} (0.595) & 2.788 (0.273) & \textbf{2.492} (0.484)\\ \bottomrule
\end{tabular}
\caption{Evaluation results ($\pm$ standard deviation) on GT set. \emph{Rel.} and \emph{Coh.} are short for relevance and coherence.}
\label{tab:gold_results}
\end{center}
\end{table*}

In this work, we crawled 5,447 MWPs of linear equations from a third-party website, and each MWP consists of two unknown variables and two equations. It covers 119 topics and the average length of an MWP is 62 words. In one CSKG, the average number of entities is 17.067 and the average number of edges is 29.102. We randomly select 544 of them as our validation set, and 546 of them as our gold-standard test (GT) set. Please note that different from previous work of automatically solving the MWPs such as \emph{MAWPS} \citep{koncel2016mawps} and \emph{MathQA} \citep{amini2019mathqa}, we focus on the generation task of MWPs of more than one linear equations in students' real-life scenarios by using topics in our CSKG. Both \emph{MAWPS} and \emph{MathQA} datasets do not contain MWPs that have two or three equations and variables. Furthermore, there is no explicit topics associated with the MWPs in these publicly available MWP datasets.

We use following evaluation metrics: (1) \emph{BLEU-4}: the 4-gram overlap score against gold-standard sentences \citep{papineni2002bleu}; (2) \emph{METEOR}: n-gram overlap with paraphrase and language-specific considerations \citep{denkowski2014meteor}; (3) \emph{ROUGE-L}: the overlap of longest common subsequence between candidate and gold-standard sentences \citep{lin2004rouge}; (4) \emph{Self-BLEU}: the diversity measurement of averaging BLEU scores of four generated MWP pairs given the same input \citep{zhu2018texygen}.


Meanwhile, we conduct two human evaluation studies to comprehensively evaluate the quality of the generated MWPs. First, we ask three evaluators to rate from the following aspects ranging from 1 to 3: (1) \emph{Equation Relevance}: how relevant is MWP with respect to the input equations? (2) \emph{Topic Relevance}: how relevant is MWP with respect to the given topic? and (3) \emph{Language Coherence}: whether the MWP is coherent and well-organized. We use the average scores from three human evaluators as our final results.

Before training, in order to ensure that each question is answerable, we first use sympy's equation solver\footnote{https://docs.sympy.org/latest/modules/solvers/solvers.html} to solve all the algebraic equations in the dataset. We tokenize our training data with BPE method \citep{kudo2018sentencepiece} and extend the subword vocabulary with our pre-defined special tokens. More preprocessing details can be found in Appendix \ref{sec:preprocessing}. During training, we initialize the GGNN parameters with normal distribution $N(0,0.02)$ and the number of GGNN hops is set to 3. The dimension of word embedding is 128 with random initialization. We utilize GRU for all RNNs and the hidden state size is 512. The size of sampled latent variable is set to 128, and we apply reparameterization trick during training and inference. We set the teacher forcing probability to 0.5 and train our model using Adam optimizer \citep{kingma2014adam} with learning rate scheduling. The batch size is set to 32 and the beam search width is set to 5. All hyper-parameters are tuned on the development set. We use linear KL annealing technique following \citet{fu2019cyclical} to alleviate the KL collapse problem and apply scheduled sampling to alleviate the exposure bias problem in GRU training \citep{bengio2015scheduled}.



We compare our \emph{MaKE} against several strong baselines: (1) the template based method, i.e., \emph{Template}; (2) conditional VAE that captures the diversity in the encoder and uses latent variables to learn a distribution over potential intents, i.e., \textit{CVAE} \citep{zhao2017learning}; (3) an RNN-based seq2seq model with equation-topic fusion mechanism and entity-enforced loss, i.e., \emph{MAGNET} \citep{zhou2019towards}; (4) the SOTA pre-trained language model with a shared Transformer network and self-attention masks, i.e., \textit{UniLM} \citep{dong2019unified}; and (5) a standard Transformer-based seq2seq model, i.e., \textit{Transformer} \citep{Vaswani2017attention}. More details are provided in Appendix \ref{sec:baseline-appendix}.










Please note that we do not select rewriting based approaches as the baselines in this work. This is because rewriting based approaches require a very large pre-stored question bank and it only works when the input equations are matched in the question bank.
\begin{table*}[!hbtp]
\footnotesize
\begin{center}
\begin{tabular}{lp{14cm}lp{0.5cm}}
\toprule
\multicolumn{2}{l}{\multirow{1}{*}{\textbf{Equations}: y-x=6; 8y-4x=64; \quad \textbf{Topic}: Rowing boat; \quad \multirow{1}{*}{\textbf{Entities}: x: Small boat; \quad y: Big boat;}}} \\
\midrule
CVAE & There are 6 in ChangLong Park. \textcolor{red}{<unk> 4 small boats, 8 big boats, small boat can carry 64 people. There are 64 more people picking in big boats} than in small boats. How many people are there on the big boat?\\
\midrule
MAGNET & \textcolor{red}{Reward}, small boat, each can sit 4. Big boat, each can sit 8. \textcolor{red}{Small boat more than 6, more than 64}. Please tell the number of small boat?\\
\midrule
UniLM & There are small boats and big boats in the competition. There are 6 sitting in big boats, \textcolor{red}{8 big boats.} On the scene, \textcolor{red}{one is more than the big boat with 8 people,} there are sitting in small boat,  \textcolor{red}{64 people for total.}\\
\midrule
Transformer & In order to reward the students who did well in this test, Teacher Fang decided to \textcolor{red}{take 4 and 6 with a total of 64 people} to go boating on the weekend! A small boat can seat 4 people, and a big boat can seat 8 people. Teacher Fang rents \textcolor{red}{6 more small boats than the big boats.} How many small boats does Teacher Fang rent?\\
\midrule
MaKE & A company has two types of boats, and the number of big boats is 6 more than that of small boats. Each small boat can accommodate 4 people and each big boat can accommodate 8 people. When all boats are filled up, the number of people in the small boats is 64 less than that in the big boats. How many big boats are there?\\
\bottomrule
\end{tabular}
\caption{Illustrative examples of the MWP generation comparison with unseen equations. There is no results from \textit{Template} because it doesn't work on unseen equations. The incorrect part is highlighted in red color.} 
\label{tab:unsee_example}
\vspace{-0.6cm}
\end{center}
\end{table*}

\subsection{Results and Analysis}
\label{sec:results}

\textbf{Evaluation Results on GT Set.} Results on the GT set are listed in Table \ref{tab:gold_results}, which shows that our \textit{MaKE} outperforms all baseline methods in terms of both automatic and human evaluation metrics. Specifically, from Table \ref{tab:gold_results}, we find: (1) comparing \textit{MaKE} and \textit{Template}, \textit{Template} doesn't perform well in language coherence and topic relevance. This is because the MWP templates are stereotyped. Mismatches between the template context and the re-filled words lead to incoherent texts; and (2) comparing \textit{MaKE} and seq2seq baselines, with rich representations of equations and CSKG, \textit{MaKE} is able to better capture mathematical relations and improve MWP quality with real-life plots under the given topic.

\noindent \textbf{Turing Test Results on GT Set.} For each existing MWP in the GT set, we generate a new MWP of the same equations but with a different topic. We show such pairs to the human evaluators and ask them to distinguish which one is the generated MWP. We measure the results of this artificial ``Turing Test'' via \emph{Fool Ratio}, i.e., the fraction of instances in which a model is capable of fooling the evaluators. Ideally, perfect MWP generation will lead to random guesses and the ideal \emph{Fool Ratio} would be 50\%. Finally, we get an averaged \emph{Fool Ratio} of 39.38\% (36.08\%, 42.49\% and 39.56\% from three annotators respectively). This demonstrates that the generation quality is 78.76\% (39.38/50) as good as the quality from human teachers.


\noindent \textbf{Ablation Study.} We compare MaKE with three different ablation methods, namely \textit{MaKE w/o symbolic}, \textit{MaKE w/o CSKG} and \textit{MaKE w/o planning}. Specifically, for \textit{MaKE w/o symbolic}, we only input the CSKG to the model, get rid of the equation-based symbolic graph and leave the other components unchanged. For \textit{MaKE w/o CSKG}, we retain symbolic graph in the input and discard CSKG. For \textit{MaKE w/o planning}, the input remains unchanged, but the decoder becomes a normal GRU and the attention score is computed for all nodes in symbolic graph and CSKG simultaneously.

Table \ref{tab:gold_results} shows the results of ablation study. Without the self-planning module, we observe that the model's self-BLEU performance has decreased, which empirically supports our assumption that the design of self-planning module can capture the flexibility in the language. Meanwhile, the performance of our model drops by 1.71\% in \textit{BLEU-4}, 0.75\% in \textit{METEOR}, 0.02\% in \textit{ROUGE-L} and 2.259\% in \textit{Self-BLEU}, which also proves the effectiveness of self-planning module. The \textit{MaKE w/o CSKG} approach achieves the best \textit{Self-BLEU} score but the worst human evaluation scores, which indicates that the representations of CSKG help form valid MWPs in real-life scenarios. This is because we  utilize CSKG as a commonsense constraint on the generated MWPs, which results in a limited number of words that can be generated by the \textit{MaKE} method under that condition. When we remove such constraint in \textit{MaKE w/o CSKG}, the model only needs to satisfy the symbolic equation conditions, regardless of what the topic is or which entity the unknown variable corresponds to. Hence the search space for words will become larger, which will directly increase the Self-BLEU score. However, it has the drawback that may cause the generated texts to violate the commonsense knowledge. The \textit{MaKE w/o symbolic} shows a significant decrease on all the automatic evaluation metrics except for the topic relevance score, which is reasonable since understanding of the mathematical variables, numerical quantities, operations, and their relations is essential in generating logical coherent MWPs.



\begin{table}[!htbp]
\footnotesize
\begin{center}
\begin{tabular}{p{7.6cm}}\toprule
\textbf{Equations}: x=y; 2x+4y=48; \quad \textbf{Topic}: Livestock\\
\midrule
\textbf{Entities}: x: Chicken; \quad y: Rabbit;\\
\midrule
1. Rabbits and chicken are in one cage. The number of rabbits is 0 less than that of chickens. They have 48 legs in total. How many rabbits and chickens in cage?\\ 
\midrule
2. There are the same number of chickens and rabbits in the yard, and the total number of legs is 48. How many rabbits and chickens are in the yard?\\
\midrule
3. Chicken and rabbits are in the same cage. Xiaoming counted the number of heads of the two animals and found that the number of chicken heads was 0 more than the number of rabbit heads. There are 48 legs in total. May I ask how many chickens are there?\\
\midrule
4. Xiaojun is very good at math, but today there is a difficult problem for him: A chicken has 1 head and 2 legs, and a rabbit has 1 head and 4 legs. There are chickens and rabbits in the same cage, and the number of chickens is equal to the number of rabbits. There are 48 feet in total, so how many chickens and how many rabbits are there? \\
\bottomrule
\end{tabular}
\caption{An illustrative example of the diverse MWP generation made by \textit{MaKE}.}
\label{tab:case-diversity}
\end{center}
\vspace{-0.4cm}
\end{table}

\noindent \textbf{Qualitative Case Study.} Because of the GGNN encodings of equations, our \textit{MaKE} model is able to handle a wide range of mathematical relations, including both addition and subtraction, i.e., $a$, $b$, $m$, $c$, $d$ and $n$ may be either positive or negative in $ax + by = m; cx + dy = n$. We quantitatively compare the generation quality of \textit{MaKE} with other baselines and the results are shown in Table \ref{tab:unsee_example}. Furthermore, we show the diverse results of \textit{MaKE} qualitatively in Table \ref{tab:case-diversity}. Additional examples can be found in Appendices \ref{sec:additional_unseen} - \ref{sec:diverse}. As we can see, (1) \textit{CVAE} and \textit{Transformer} cannot interpret the equations correctly and fail to generate desired MWPs; (2) our \textit{MaKE} approach is able to generate diverse enough MWPs in real-life scenarios.


\noindent \textbf{Large-scale Human Evaluation Results.} Besides evaluations on the GT set, which is usually limited in educational scenarios \citep{xu2019learning,wang2020representation}, we conduct evaluations on the large-scale generated results. We randomly create 100 valid linear equations and ensure that none of them appears in our training set. Meanwhile, we select top 30 common real-life topics. For each pair of equation and topic, we generate 5 MWPs accordingly and therefore, we obtain 15,000 MWPs. We conduct a human evaluation to assess the quality of these generated MWPs and the results are shown in Table \ref{tab:unseen-test}. We can see that our method outperforms baseline models by a large margin. 

\begin{table}[!hbtp]
\footnotesize
\center
\scalebox{0.95}{
\begin{tabular}{@{}lccc@{}} \toprule
Method & Equation Rel. & Topic Rel. & Language Coh.\\ 
\midrule
CVAE           & 1.583 (0.493) & 2.550 (0.487) & 1.366 (0.605) \\
MAGNET & 1.603 (0.430) & 2.467 (0.222) & 1.517 (0.508) \\
UniLM          & 1.451 (0.530) & 2.416 (0.690) & 1.150 (0.441) \\
Transformer    & 1.699 (0.395) & 2.351 (0.391) & 2.416 (0.460) \\
\midrule
MaKE & \textbf{2.308 (0.507)} & \textbf{2.558 (0.228)} & \textbf{2.505 (0.461)}\\ \bottomrule
\end{tabular}
}
\caption{Evaluation results ($\pm$ standard deviation) on the large-scale generated MWP data. There is no results from \textit{Template} because it doesn't work on unseen equations.}
\label{tab:unseen-test}
\end{table} 

\noindent \textbf{Error Analysis.} To better understand the limitation of our approach, we manually review 150 equations and the corresponding generated MWPs. The two major problems are: missing information and language disfluency. We show two representative examples in Table \ref{tab:error_example}. In the example of missing information, the information that the small boat can accommodate 2 people and big boat can accommodate 4 people are ignored because some MWPs in the training set often ignore this ``pre-leaned'' knowledge like chickens have two legs. Language disfluency problem is introduced due to the limit size of training data under certain specific topic. This can be alleviated or addressed by either collecting more MWP training data or provide more information in CSKG to explicitly control the context of the generated text, such as the fact that livestock often live on farms and marine animals are found in the ocean, etc.

\begin{table}[!hbtp]
\footnotesize
\begin{center}
\begin{tabular}{l|p{4.5cm}lp{0.5cm}}
\toprule
\multicolumn{2}{l}{\multirow{1}{*}{\textbf{Equations}: x-y=6; \textcolor{red}{2}x-\textcolor{red}{4}y=10; \quad \textbf{Topic}: Rowing boat}} \\
\midrule
\multicolumn{2}{l}{\multirow{1}{*}{\textbf{Entities}: x: Small boat; y: Big boat;}}\\
\midrule
Missing information & Teacher Mr.Huang and his 35 students come to row the boat. They find 6 more small boats than big boats. There are 10 more students in the small boat than in the big boat. How many big boats are there? \\
\midrule
\multicolumn{2}{l}{\multirow{1}{*}{\textbf{Equations}: x-y=1; 6x-8y=0; \quad \textbf{Topic}: Insects}} \\
\midrule
\multicolumn{2}{l}{\multirow{1}{*}{\textbf{Entities}: x: Cockroaches; y: Ants;}}\\
\midrule
Language disfluency & There are two types of heads: cockroaches and ants. Cockroaches have 1 more head than ants, and cockroaches have \textcolor{red}{0 more than ant legs}. How many cockroaches and ants respectively? \\
\bottomrule
\end{tabular}
\caption{Illustrative examples that demonstrate the typical problems of the current system.}
\label{tab:error_example}
\end{center}

\end{table}

\vspace{-0.5cm}
\section{Conclusion}
\label{sec:conclusion}
In this paper, we presented a neural encoding-decoding architecture for MWP generation. Comparing with the existing NQG algorithms, the advantages of our \emph{MaKE} are: (1) it extracts intrinsic representations of both the equation based symbolic graph and the CSKG; (2) it automatically selects and incorporates information from equations and knowledge graphs during the decoding process; and (3) it is able to generate relevant, coherent and diverse MWPs in students' real-life scenarios. Experimental results on real-world educational MWP data sets demonstrate that \emph{MaKE} outperforms other SOTA NQG approaches in terms of both automatic evaluation metrics and human evaluation metrics. In the future, we plan to explore the MWP generation problems for more mathematical variables with high-order operations, and explore the method to incorporate commonsense knowledge from publicly available CSKG like ConceptNet or Wikipedia.

\section*{Acknowledgements}
This work was supported in part by National Key R\&D Program of China, under Grant No. 2020AAA0104500 and in part by Beijing Nova Program (Z201100006820068) from Beijing Municipal Science \& Technology Commission.

\bibliographystyle{acl_natbib}
\bibliography{emnlp2021}

\clearpage
\appendix
\section{Appendix}
\subsection{Equation Variants}
\label{sec:equation_variants}
In our scenario, the general expression formula can be formed as follows: 

\begin{align}
\eta_0 \varphi_0 \emph{x} \eta_1 \varphi_1 \emph{y} &= \varphi_2 \eta_2 \varphi_3 \label{eq:eq1}\\
\eta_3 \varphi_4 \emph{x} \eta_4 \varphi_5 \emph{y} &= \varphi_6 \eta_5 \varphi_7  \label{eq:eq2}
\end{align}

\noindent where $\eta_* \in \{+, -, \times, \div\}$ are operators in equations, and $\varphi_*$ are numeric numbers. To be noticed, only one of the operators between $\eta_0$ ($\eta_3$) and $\eta_1$ ($\eta_4$) may be a minus operator, or neither. The equation variants are derived from different configurations of operators $\eta_*$.

\begin{figure}[!hbpt]
    \centering
    \includegraphics[width=\linewidth, scale=1.00]{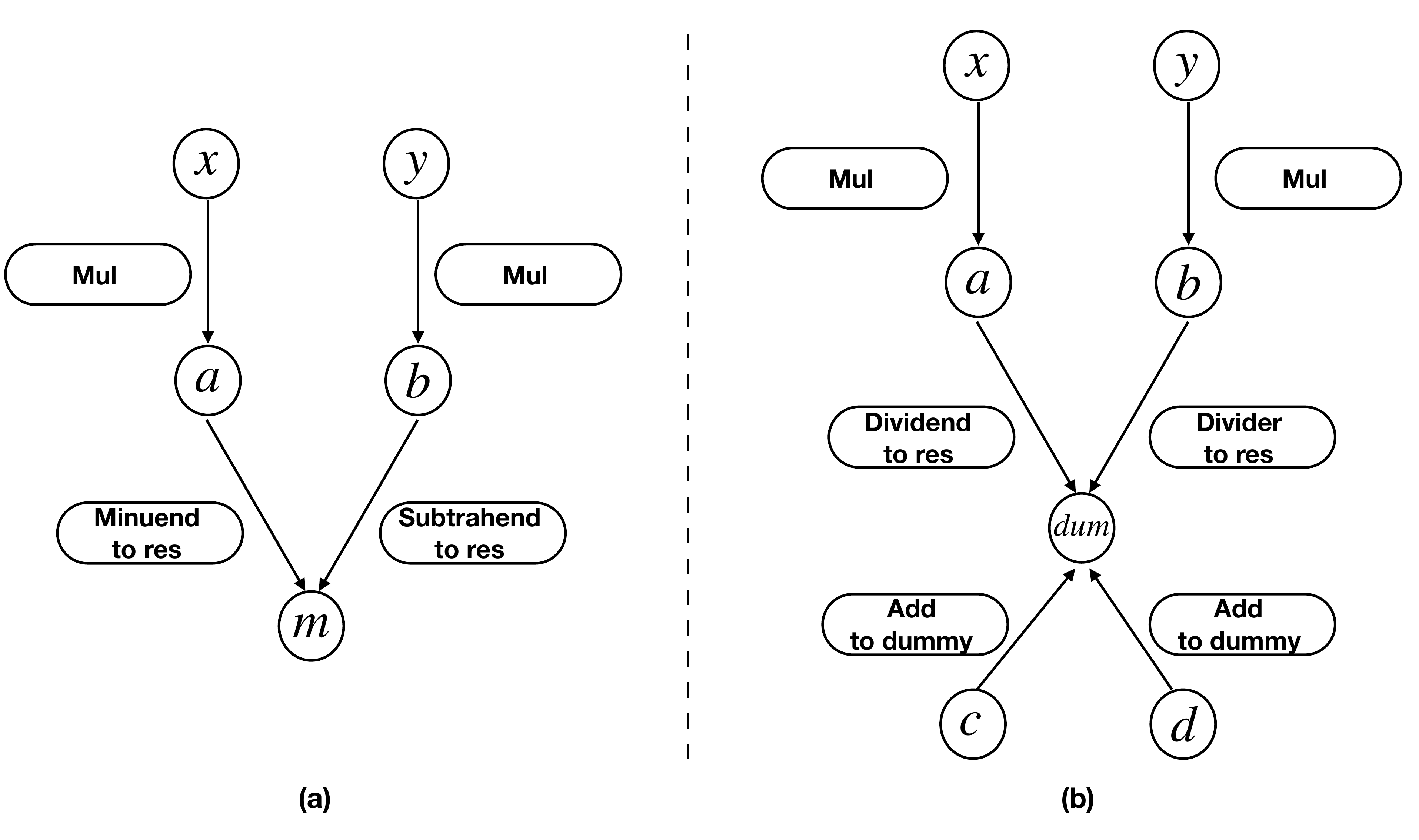}
    \caption{(a) a symbolic graph of equation ``ax-by=m''; (b) a symbolic graph of equation ``ax/by=c+d''; \emph{dum} denotes the dummy node in the symbolic graph.}
    \label{fig:equation_variants}
\end{figure}

We go into detail for eq.(\ref{eq:eq1}) discussing all its possible variants, eq.(\ref{eq:eq2}) keeps the same behavior as eq.(\ref{eq:eq1}), and different combinations of eq.(\ref{eq:eq1}) and eq.(\ref{eq:eq2}) will lead to different system of linear equation in two unknowns. According to the presence or absence of the operator $\eta_2$ and numeric number $\varphi_3$, we show two different possible symbolic graph structures in Figure \ref{fig:equation_variants}. In Figure \ref{fig:equation_variants} (a), $\varphi_2$ is equal to ``m'', $\eta_2$ and $\varphi_3$ are empty. Thus we connect node $a$ and node $m$ with relation ``Minuend to res''; connect node $b$ and node $m$ with relation ``Subtrahend to res''; representing $ax$ and $by$ are  minuend and subtrahend element in ``ax-by=m'' respectively. In Figure \ref{fig:equation_variants} (b), $\varphi_2$, $\eta_2$ and $\varphi_3$ are equal to ``c'', ``+'' and ``d'' respectively. In order to be consistent with the graph structure described in Figure \ref{fig:equation_variants} (a), we first add a dummy node $dum$ in our symbolic graph, then connect node $c$ and node $dum$ with relation ``Add to dummy'', and connect node $d$ and node $dum$ with relation ``Add to dummy''. In this way, the dummy node can represent the expression ``c+d''.

\begin{table*}[!htpt]
\begin{center}
\begin{tabular}{ll}
\hline
\multicolumn{1}{l}{\multirow{2}{*}[1ex]{\textbf{Equation templates}: x+y=\emph{$\alpha$}; bx-cy=d}} & \multicolumn{1}{l}{\multirow{2}{*}[1ex]{\textbf{Topic}: vehicle}}\\
\hline
\multicolumn{2}{l}{\textbf{Query text}: There are \emph{$\alpha$} \emph{x\_entity} and \emph{y\_entity} in the parking lot. Each \emph{x\_entity} has \emph{b} wheels} \\
\multicolumn{2}{l}{and each \emph{y\_entity} has \emph{c} wheels. \emph{x\_entity} has \emph{d} more total wheels than \emph{y\_entity}. How many}\\
\multicolumn{2}{l}{\emph{x\_entity} are there?}\\
\hline
\multicolumn{2}{l}{\textbf{Query variables:} \emph{x\_entity}: motorcycles, \emph{y\_entity}: cars}\\
\hline
\multicolumn{2}{l}{\textbf{Generated MWP} There are 6 motorcycles and cars in the parking lot. Each motorcycles has} \\
\multicolumn{2}{l}{2 wheels and each cars has 4 wheels. Motorcycles has 6 more total wheels than cars. How m-} \\
\multicolumn{2}{l}{-any motorcycles are there?}\\
\hline
\end{tabular}
\caption{Query process in our template-based-method.}
\label{tab:template-based-method}
\end{center}
\end{table*}

\begin{table*}[!htpt]
\begin{center}
\begin{tabular}{ll}
\hline
\multicolumn{1}{l}{\multirow{2}{*}[1ex]{\textbf{Input equation}: x+y=6; 2x-4y=6}} & \multicolumn{1}{l}{\multirow{2}{*}[1ex]{\textbf{Topic}: vehicle}}\\
\hline
\multicolumn{1}{l}{\textbf{General expression formula}:} & \multicolumn{1}{l}{$\eta_0 \varphi_0 \emph{x} \eta_1 \varphi_1 \emph{y} = \varphi_2 \eta_2 \varphi_3$}\\
& $\eta_3 \varphi_4 \emph{x} \eta_4 \varphi_5 \emph{y} = \varphi_6 \eta_5 \varphi_7$ \\
\hline
\multicolumn{2}{l}{\textbf{Input sequence expression}: [$\eta_0, \varphi_0, \eta_1, \varphi_1, \varphi_2,  \eta_2, \varphi_3, \eta_3, \varphi_4, \eta_4, \varphi_5, \varphi_6, \eta_5, \varphi_7$, Topic]} \\
\hline
\multicolumn{2}{l}{\textbf{Input sequence for given example}: [pad,  $\varphi_0$, +, $\varphi_1$, $\varphi_2$,  pad, pad, pad,  $\varphi_4$, -, $\varphi_5$, $\varphi_6$,} \\
& pad, pad, \emph{x\_entity}, \emph{y\_entity},vehicle] \\
\hline
\multicolumn{2}{l}{\textbf{Output MWP}: There are $\varphi_2$ \emph{x\_entity} and \emph{y\_entity} in the parking lot. Each \emph{x\_entity} has $\varphi_4$} \\
\multicolumn{2}{l}{wheels and each \emph{y\_entity} has $\varphi_5$ wheels. \emph{x\_entity} has $\varphi_6$ more total wheels than \emph{y\_entity}. How } \\
\multicolumn{2}{l}{many \emph{x\_entity} are there?}\\
\hline
\end{tabular}
\caption{Input sequence for seq2seq method, $\varphi_*$ are numeric number in equations, and $\eta_*$ are operators. If there is no valid operator or number for a given special token, we fill it with a pad token.}
\label{tab:seq2seq-method}
\end{center}
\end{table*}

\subsection{Levi Graph Transformation}
\label{sec:levi_transformation}

Let $\mathcal{G} = \{\mathcal{V}, \mathcal{E}, \mathcal{R}\}$ be a directed symbolic graph  with nodes $v_i \in \mathcal{V}$ and labeled edges $(v_i, r, v_j) \in \mathcal{E}$. As shown in Figure \ref{fig:dual_graph} (a), where $r \in \mathcal{R}$ is a relation type, i.e., \emph{Add to res}, \emph{Mul}, etc. Let $|\mathcal{V}|$ and $|\mathcal{E}|$ denote the number of nodes and edges, respectively. We convert the graph $\mathcal{G}$ into an unlabeled and directed bipartite graph $\mathcal{G}^t=\{\mathcal{V}^t, \mathcal{E}^t\}$ with levi transformation by converting each labeled edge $(v_i, r, v_j) \in \mathcal{E}$ into two unlabeled edges $(v_i, r), (r, v_j) \in \mathcal{E}^t$, where $|\mathcal{V}^t| = |\mathcal{V}| + |\mathcal{E}|$. Intuitively, transforming a graph into its Levi graph form turns original edges into additional nodes, which allows us to directly encode edge label information with word embeddings and guarantee the relation message passing with multi-hop reasoning.

\subsection{Training and Testing}
\label{sec:preprocessing}

We obtain non-lexical text by replacing the numbers in the question text with the pre-defined special tokens in our symbolic equation graph and CSKG. The procedures are similar to the example in Table \ref{tab:seq2seq-method} with the following differences: 
\begin{itemize}
    \item Matching words for unknown variables are first extracted from the gold MWP, and query our private database with the given topic word to construct our commonsense knowledge graph.
    \item \textit{MaKE} transforms operators $\eta_*$ into equation graph edge labels (relations), and numeric number $\varphi_*$ into equation graph nodes $v$.
    \item More words in MWP texts are replaced with special tokens in CSKG. Take the sentence in Table \ref{tab:seq2seq-method} as an example, \emph{wheels} are replaced by one node (\emph{counting entity}) in the corresponding commonsense graph.
\end{itemize}

During training \textit{MaKE}, the input is CSKG and the equation based symbolic graph, and the output target is the delexicalized words sequence. We apply the same word refilling post-processing procedure to obtain the final MWP.



\subsection{Baseline Methods Details}
\label{sec:baseline-appendix}
\textit{Template} In addition to neural baselines, we use a problem-specific, template-based generator. The template-based method first finds MWP problems with the same type of input equations in the question bank given the input topic words. For instance, the query equation is \emph{$x+y=6; 2x-4y=6$} and the query topic is \emph{vehicle}. As shown in Table \ref{tab:template-based-method}, we first delexicalize the input equation pairs with special tokens and save them for post-processing. After query our question bank with the delexicalized equation pairs and the topic word, we obtain the pre-stored MWP template and matching words for unknown variables. Finally we fill the MWP template with the previously saved delexicalized words and obtain the generated MWP.

\textit{CVAE} \citep{zhao2017learning} Similar to previous work, we apply a seq2seq model and adopt a latent variable to capture the diversity of MWPs. We replace the hierarchical encoder with a one-layer GRU, and the initial state of the decoder is the combination of a latent variable and the final state of the encoder. As shown in Table \ref{tab:seq2seq-method}, we apply the delexicalization process and sequence transformation operations for all the training data. The input sequence includes special tokens, operators and the topic word. After refilling the special tokens with corresponding matching words, we obtain the final MWP.

\textit{MAGNET} \citep{zhou2019towards} MAGNET is a previously proposed seq2seq MWP generation framework. Following the original implementation, we utilize a bidirectional RNN to encode equation sequence and encode topic word with a word representation lookup table. The decoder is a single directional RNN with equation-topic fusion mechanism to leverage both equation and topic information. We follow the same input sequence as described in Table \ref{tab:seq2seq-method}, but split the equation sequence and the topic words as separate inputs.

\textit{UniLM} \citep{dong2019unified} A pre-trained natural language generation model with transformer encoder and decoder blocks. We fine-tune UNILM on MWP generation task with the same input and output token sequence described in CVAE method.

\textit{Transformer} \citep{Vaswani2017attention} We included a Transformer-based seq2seq model which has proved its success in machine translation tasks. The same input sequence as described in previous method.

\subsection{Additional MWP Generation Comparison}
\label{sec:additional_unseen}

We provide additional illustrative examples of the MWP generation comparison with unseen equations in Table \ref{tab:unsee_appendix}.

\begin{table*}[!htbp]
\footnotesize
\begin{center}
\begin{tabular}{lp{11.5cm}lp{1cm}}
\toprule
\multicolumn{2}{l}{\multirow{1}{*}{\textbf{Equations}: x=2y; 4x+6y=56; \quad \textbf{Topic}: Vehicle}} \\
\midrule
\multicolumn{2}{l}{\multirow{1}{*}{\textbf{Entities}: x: Small car; \quad y: Large car;}}\\
\midrule
\multicolumn{1}{l}{\textit{CVAE}} & There are many small cars and big cars in the parking lot. There are \textcolor{red}{6 people in these cars} and \textcolor{red}{56 people in big cars}. How many small cars and big cars are there? \\
\midrule
\textit{MAGNET} & We need to clean up a total of large cars and small cars. We know that \textcolor{red}{small car can transport 6}, and \textcolor{red}{the number of large cars is the number of times}. Please tell the number of small cars?\\
\midrule
\textit{UniLM} & Doctors have produced a lot of small cars and large cars. There are \textcolor{red}{0 cars in total}. These two types of cars have 56 people. How many of these two cars \textcolor{red}{have a total of 56 people, small cars?} \\
\midrule
\textit{Transformer} & There are 56 people in the class to \textcolor{red}{go rowing}, and there are \textcolor{red}{0 cars in total}. Among them, the small cars have 4 people each, and the big cars have 6 people each. The small cars have $( )$, and the big cars have $( )$.\\
\midrule
\textit{MaKE} & A company needs to transport 56 wheels in one transportation task. We know that each large car can carry 6 wheels and each small car can carry 4 wheels. The company has completed the task with only one time transportation, and the number of used small car is twice the number of large car. Q: They used () small car. \\
\bottomrule
\end{tabular}
\begin{tabular}{lp{11.5cm}lp{1cm}}
\toprule
\multicolumn{2}{l}{\multirow{1}{*}{\textbf{Equations}: -x+y=20; 2x-4y=10; \quad \textbf{Topic}: Livestock}} \\
\midrule
\multicolumn{2}{l}{\multirow{1}{*}{\textbf{Entities}: x: Chicken; \quad y: Rabbit;}}\\
\midrule
\multicolumn{1}{l}{\textit{CVAE}} & Chickens and rabbits in the same cage. The number of chickens is \textcolor{red}{twice the number of chickens}. The number of rabbits and rabbits is \textcolor{red}{twice the number of chickens}, and the \textcolor{red}{total number of legs is 10}. How many chickens and rabbits are there each? \\
\midrule
\textit{MAGNET} & Chickens and rabbits in the same cage. \textcolor{red}{The number of chickens is 20 more than that, and there are 10 rabbit legs}. How many rabbits are there?\\
\midrule
\textit{UniLM} & How many chickens and rabbits are there in the \textcolor{red}{10-legged training centre}? \\
\midrule
\textit{Transformer} & There were \textcolor{red}{20 chickens and rabbits}. The total number of legs of the chickens was 10 more than the rabbits, and the chickens had $( )$ only.\\
\midrule
\textit{MaKE} & The number of chickens is 20 less than the number of rabbits, and the number of feet for chickens is 10 more than that of rabbits. How many chickens and rabbits are there? \\
\bottomrule
\end{tabular}
\begin{tabular}{lp{11.5cm}lp{1cm}}
\toprule
\multicolumn{2}{l}{\multirow{1}{*}{\textbf{Equations}: x+y=192/16; 22x+14y=192; \quad \textbf{Topic}: Dormitory}} \\
\midrule
\multicolumn{2}{l}{\multirow{1}{*}{\textbf{Entities}: x: Large dormitory; \quad y: Small dormitory;}}\\
\midrule
\multicolumn{1}{l}{\textit{CVAE}} & 
\textcolor{red}{In the average daily None dormitory of the fifth grade, Teacher Chen gave four students}, two small dormitories can accommodate 22 students. \textcolor{red}{Two groups can accommodate 22 students}, and the residence is just so big that the dormitory can accommodate 192 students. There are () rooms in the dormitory group. There are 14 students in the small dormitory, which just happened to be full. \\
\midrule
\textit{MAGNET} & \textcolor{red}{There are 16 people in HongKong}. The large dormitory can accommodate 22 people, and the small dormitory can accommodate 14 people. A total of 192 people, how many large dormitory?\\
\midrule
\textit{UniLM} & The school arranges accommodation for students. The big dormitory can accommodate 22 students in the dormitory, and the small dormitory can accommodate students. A total of 192 students can live in the dormitory. On average, \textcolor{red}{how many rooms are there in each small dormitory}? \\
\midrule
\textit{Transformer} & \textcolor{red}{192 students from grade 1 to grade 6 go to the spring trip. There are 16 teachers in total.}  Students have two kinds of dormitories to choose. The large dormitory can live in 22 people, and the small dormitory can live in 14 people. In total, 192 dormitories are booked. What is the number for the booked small dormitory and large dormitory separately? \\
\midrule
\textit{MaKE} & The Youth Hostel is designed to accommodate 22 people in the large dormitory and 14 people in the small dormitory. One day, there were 192 travelers and the hostel was just about full with an average of 16 people per dormitory. How many large dormitories are there in the youth hostel?\\
\bottomrule
\end{tabular}    
\caption{Illustrative examples of the MWP generation comparison with unseen equations. $( )$ represents the question that the student needs to solve.}
\label{tab:unsee_appendix}
\end{center}
\end{table*}

\subsection{Additional Diverse MWP Results}
\label{sec:diverse}

We provide additional diverse MWP results in Table \ref{tab:appendix_diversity}.

\begin{table*}
\footnotesize
\begin{center}
\begin{tabular}{p{13cm}}\toprule
\textbf{Equations}: x+y=100; 20x+30y=2600; \quad \textbf{Topic}: Buy ticket\\
\midrule
\textbf{Entities}: x: front row ticket; y: back row ticket\\
\vspace{0.1cm}
1. The company organized staff to go to the Shanghai Circus to see the show. Tickets purchased on the official website would receive discounts, 20 yuan off for each front row ticket and 30 yuan off for each back row ticket. They bought 100 tickets on official website and saved 2,600 yuan, how many front row tickets did they buy?\\
\midrule
\textbf{Entities}: x: child ticket; y: adult ticket\\
\vspace{0.1cm}
2. There are adult and child tickets available at the acrobatic ticket office in Linjiang Park. The price of the adult ticket is 30 yuan and the price of the child ticket is 20 yuan. A total of 100 acrobatic tickets were sold today, and the revenue was 2,600 yuan. How many adult tickets were sold?\\
\midrule 
\textbf{Entities}: x: one-way ticket; y: round-trip ticket\\
\vspace{0.1cm}
3. The bus station sold 100 tickets today and received 2,600 yuan. There were two kinds of tickets on their website, a round-trip ticket for ¥30 and a one-way ticket for ¥20. How many round-trip tickets were sold today?\\
\bottomrule
\end{tabular}
\begin{tabular}{p{13cm}}\toprule
\textbf{Equations}: -y+x=20; 2y+4x=66; \quad \textbf{Topic}: Livestock\\
\midrule
\textbf{Entities}: x: cow; y: duck\\
\vspace{0.1cm}
1. When Sun Wukong returned to Mount Huaguo, he showed a magic spell to the monkeys. A group of cows and a group of ducks emerged. After counting, they found that there were 66 legs. If we know that the number of cows is 20 more than that of ducks. Then how many cows and ducks for each?\\ 
\midrule
\textbf{Entities}: x: cow; y: duck\\
\vspace{0.1cm}
2. Xiaohong's farm has ducks and cows. The number of ducks is 20 less than the number of cows. There are 66 legs in total, Xiaohong have $( )$ cows and $( )$ ducks.\\
\midrule
\textbf{Entities}: x: cow; y: duck\\
\vspace{0.1cm}
3. In a pasture, there are cows and ducks. There are 20 more cows than ducks, and the total number of legs of the cows and ducks is 66. There are $( )$ ducks? \\
\bottomrule
\end{tabular}
\caption{Additional illustrative example of the diverse MWP generation made by \textit{MaKE}. $( )$ represents the question that the student needs to solve.}
\label{tab:appendix_diversity}
\end{center}
\end{table*}

\end{document}